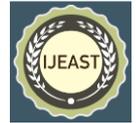

# WHITE-BOX CARTOONIZATION USING AN EXTENDED GAN FRAMEWORK

Amey Thakur
Department of Computer Engineering,
University of Mumbai,
Mumbai, MH, India

Hasan Rizvi
Department of Computer Engineering,
University of Mumbai,
Mumbai, MH, India

Mega Satish
Department of Computer Engineering,
University of Mumbai,
Mumbai, MH, India

*Abstract—* **In the present study, we propose to implement a new framework for estimating generative models via an adversarial process to extend an existing GAN framework and develop a white-box controllable image cartoonization, which can generate high-quality cartooned images/videos from real-world photos and videos. The learning purposes of our system are based on three distinct representations: surface representation, structure representation, and texture representation. The surface representation refers to the smooth surface of the images. The structure representation relates to the sparse colour blocks and compresses generic content. The texture representation shows the texture, curves, and features in cartoon images. Generative Adversarial Network (GAN) framework decomposes the images into different representations and learns from them to generate cartoon images. This decomposition makes the framework more controllable and flexible which allows users to make changes based on the required output. This approach overcomes any previous system in terms of maintaining clarity, colours, textures, shapes of images yet showing the characteristics of cartoon images.**

## I. INTRODUCTION

Cartoons are a highly popular art style that has been extensively used in a variety of contexts, from print media to children's narrative. Some cartoon artwork was created based on real-world scenes. However, manually re-creating situations from real life may be very time consuming and needs specialised abilities. Machine Learning advancements have increased the possibilities for producing visual artworks. A cartoon is a popular style of art that has been adapted for usage in a wide range of contexts. By turning real-world photos into usable cartoon scene components, the method of image cartoonization has resulted in the development of many well-known products. White box cartoonization is a method that reconstructs high-quality real-life pictures into exceptional cartoon images using the GAN framework.

The motivation of this paper is to build sophisticated cartoon animation processes that enable artists to generate material from a range of sources. Artists often use filters supplied by various software to cartoonize images/videos, however, the industry-standard software degrades the quality during translation. Additionally, meticulously recreating real-world events in cartoon styles involves a great deal of time and work, as well as a great deal of creative ability. Manually cartooning pictures frame by frame and producing high-quality cartoons is a time-consuming process. Existing software fails to produce the required outputs due to issues such as generator-discriminator stability. As a consequence, specially developed techniques for automatically converting real-world photos into high-quality cartoon-style images are very beneficial, as they enable artists to finish their work with more precision and in less time.

Many models have been developed to generate cartoon images from pictures, but have many drawbacks. CartoonGAN [1] is one of the technologies to generate cartoonized images but it adds noise and reduces the quality of an image. On the other hand, White Box Cartoonization [2] overcomes these problems and results in more precise and sharp images.

The challenges in CartoonGAN are the stability between generator and discriminator, the inaccurate positioning of the object, and understanding the perspective of images i.e. 2D or 3D as well as global objects i.e. trees, flowers, etc.

White Box Cartoonization is a variant of Black Box Cartoonization that addresses the latter's shortcomings. For example, in some cartoon workflows, by analysing the processed dataset, global palette themes are prioritised over line sharpness as a secondary concern. For others, the sharpness of objects and persons hold a great value. Black box cartoonization algorithms fail to deal with such various workflow requirements and applying a black-box model to directly fit the training data might have a severe impact on generality and stylization quality, resulting in low-quality outputs.

To get a better idea of Cartoonization, researchers consulted artists and observed cartoon painting behaviour to identify three separate cartoon image representations: a surface representation that contains smooth surfaces, a structure representation that refers to sparse colour-blocks and flattens global content in the workflow, and a texture representation that reflects high-frequency texture, contours, and details in cartoon images.

Each representation is retrieved using image processing modules, and the extracted representations are then learned and cartoonized using a generative adversarial network (GAN) architecture[3]. Tuning the value of each representation in the loss function may be used to alter the output style.

In a user survey with ten respondents and thirty pictures, the suggested technique beat three current techniques in terms of cartoon quality (similarity between the input and cartoon pictures) and quality of the product (identifying undesirable colour shifts, texture distortions, high-frequency noise or other artefacts in the images).

## II.    RELATED WORK

### A.    Preprocessing –

Preprocessing, in addition to the proposed three-step method, is a key component of our model. It aids in the smoothing of images, the filtering of features, the conversion of images to sketches, and the translation of output from one domain to another. Following the completion of these associated tasks, we can be confident that the output provided by our model will be of the highest quality.

### B.    Image to Image Translation –

For image cartoonization, we use an unpaired image-to-image [4] translation system in this article. The disadvantage of GAN is that it only works with given training data, and paired training data is not always accessible. To overcome this limitation, we use cycleGAN [5], which aims to translate a picture from a source domain $X$ to a target domain $Y$ even in the absence of paired training data. In terms of loss, we deconstruct images into many representations, forcing the network to learn various aspects with distinct goals, making the learning process more manageable and adaptable.

### C.    Generative Adversarial Network –

White Box Cartoonization uses an unsupervised learning approach. Hence we use generative modelling. In the generative model, there are samples and data i.e. x (input variable) and y (It does not have an output variable). Deep learning-based generative models are used for unsupervised learning. In short, it's a system where two networks compete with each other to create or generate variation in the data. In 2014 a framework for estimating generative networks was introduced and further many algorithms were used to enhance the adversarial process.

### D. GAN Architecture  –

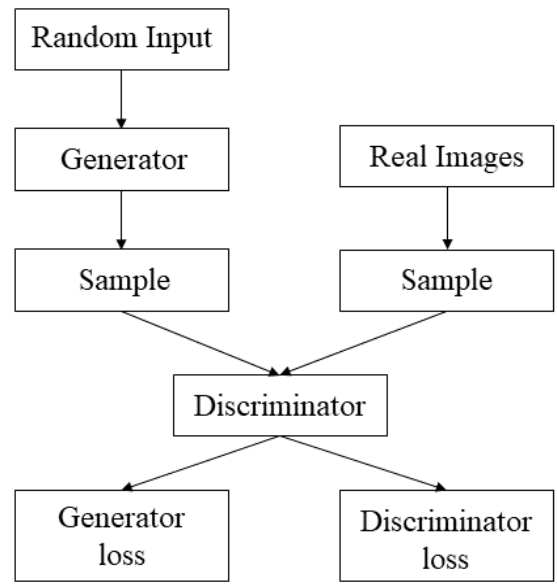

Fig. (1) Architecture of GAN

GAN is made up of two models [6], namely generative and discriminative. The generator (G) is a network that processes input and generates sample data. The discriminator (D) determines whether the data is produced or taken from the original sample by using binary classification problems and a sigmoid function that returns values between 0 and 1.

In machine learning, there are two major techniques for developing predictive models. The discriminative model is the most well-known. In this scenario, the model learns the target variable's conditional probability given the input variable, i.e. $P(Y/X=x)$. Examples include logistic regression, linear regression, and so on.

The generative model, on the other hand, learns the joint probability distribution of the input variable and the output variable, i.e. $P(X, Y) = P(X/Y) P(Y) = P(Y/X) (X)$. If the model wishes to predict something, it employs the Bayes theorem to calculate the conditional probability of the target variable given the input variable, i.e. $P(Y/X) = P(X, Y)/P (X)$.

A prominent example of generative models is Naive Bayes. The most significant advantage of the generative model over the discriminative model is that we can use it to create new instances of data because we are learning the distribution function of the data itself, which a discriminator cannot do. To generate fresh data points in GANs, we employ a generative model.

We implement the discriminator to determine whether a given data point is original or generated by our generator. Now, these two models operate in an adversarial environment, which means they compete with each other and eventually both of them improve.

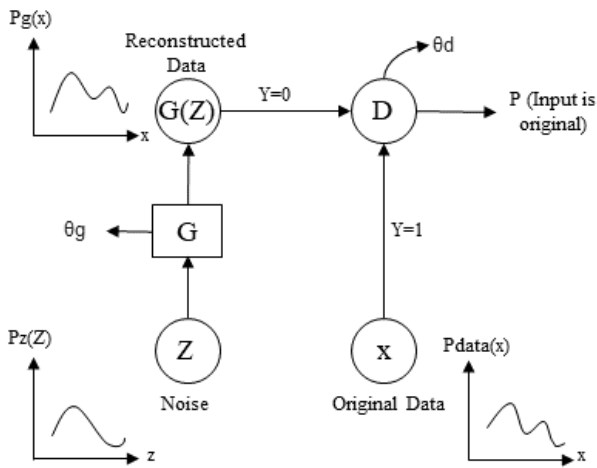

Fig. (2) High-level structure of GAN

$G$ tries to minimize the expression and $D$ tries to maximize the expression.

### III. MODEL ARCHITECTURE

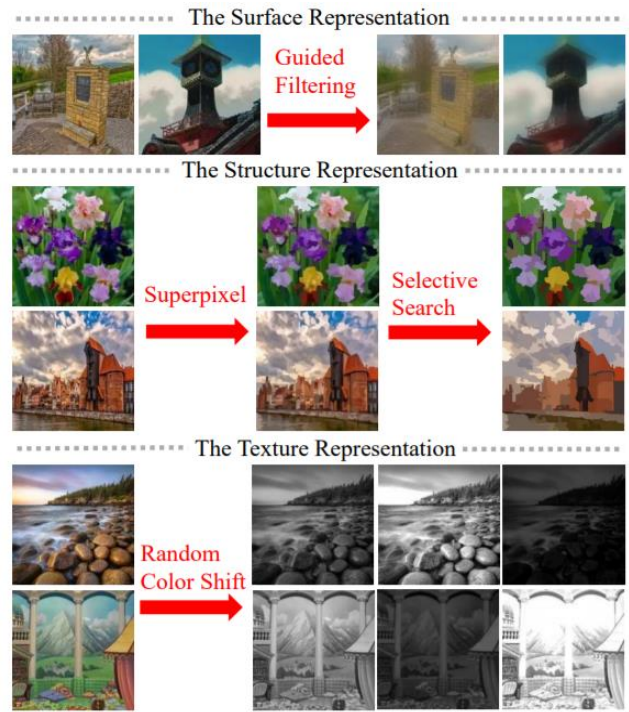

Fig. (3) Representation Techniques

$G$ and $D$ are multi-layered neural networks, $\theta g$ and $\theta d$ are the respective weights. We are using neural networks because they can approximate any function we know from the universal approximation theorem. The distribution function of the original data is shown above. In reality, we cannot really draw or mathematically compute the distribution function of original data because we input images, voices, videos and they are higher dimensional complex data so graphs are shown for mathematical analysis.

In noise distribution, we show the normal distribution because we sample randomly some data from the distribution and we feed that to our generator. $Z$ contains no information, we input it to the generator and it will produce $g$ of $z$. The domain of the original data is the same as the range of $G$ of $z$. It is necessary as we are trying to replicate original data. *Pdata* represents the probability distribution of original data. $P$ of $z$ represents the distribution of noise and $p$ of $g$ represents the distribution function of the output of the generator. We are passing original data and reconstructed data to the discriminator which will provide us with a single number that will tell the probability of the input belonging to the original data. As we can see, the discriminator is just a simple binary classifier. For training purposes, when we input original data to the discriminator we take *Y=1*. For reconstructed data we take *Y=0*. $D$ will try to maximise the chances of predicting correct classes but $G$ will try to fool $D$.

Value function for GAN:

### Value Function

$$\underset{G}{\text{Min}} \ \underset{D}{\text{Max}} \quad V(G,D) = E_x \sim P_{data}[\ln(D(x))]$$
$$+$$
$$E_z \sim P_z[\ln(1\text{-}D(G(z)))]$$

### A. Surface Representation –

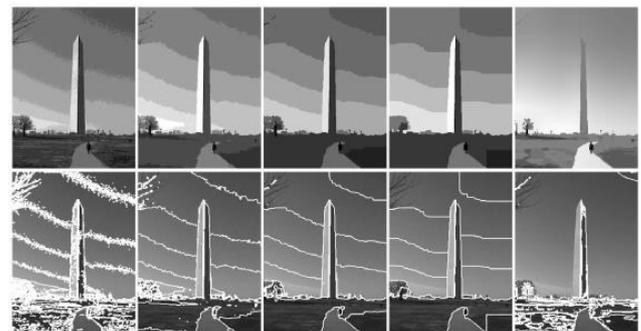

Fig. (4) 8-step edge-preserving filtering with (a)-(d) flat and (e) flexible intensity models.

Surface representation is similar to cartoon painting in which painters draw early sketches with rough strokes and possess smooth surfaces comparable to cartoon imagery. For edge-preserving filtering [7], a differentiable guided filter is used to smooth images while keeping the global semantic structure. Guided filters [8] are a more advanced variant of bilateral filters that perform better towards the edge. In simple terms, this filter preserves the edges and its information while blurring an image Examples are the median, bilateral, guided, and anisotropic diffusion filters. As compared to CartoonGAN, White box cartoonization reduces artefacts at significant levels.

### B. Structure Representation –

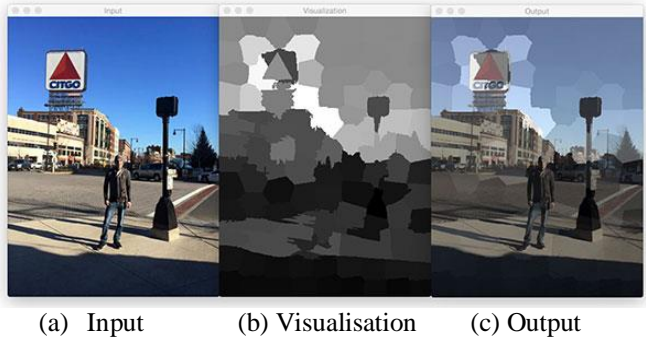

(a) Input      (b) Visualisation      (c) Output

Fig. (5)

Structure representation is used to segment images into separate regions. The segmentation is done using the Felzenszwalb algorithm [9]. This algorithm also assists us in capturing generic content information and producing outputs that are practically suitable for celluloid-style animation procedures. The pixel value average is used to colour each segmented region in standard superpixel [10] algorithms. However, they only evaluate pixel similarity and disregard semantic information; we use selective search to combine segmented regions and produce a sparse segmentation map. We discovered that this reduces global contrast, darkens images, and generates hazing effects in the final results by studying the processed dataset. To overcome these constraints, we propose an adaptive colouring algorithm. SLIC (Simple Linear Iterative Clustering) [11] algorithm is used for superpixel generation. It generates superpixels by clustering pixels on the basis of colour similarity and proximity in the image plane.

### C. Texture Representation –

The high-frequency elements of cartoon images are important learning objectives, but brightness and colour information help differentiate cartoon images from real-world photography. As a result, we suggest a random colour shift algorithm. The random colour shift can provide random intensity maps that are devoid of luminance and colour information. *Frcs (function for random colour shift)* extracts single-channel texture representation from colour images, which retains high-frequency textures and decreases the influence of colour and luminance. $Frcs\ (Irgb) = (1 − α)\ (β1 ∗ Ir + β2 ∗ Ig + β3 ∗ Ib) + α ∗ Y$ Where, *Irgb* represents 3-channel RGB colour images, *Ir, Ig* and *Ib* represent three colour channels, and *Y* represents standard grayscale image converted from RGB colour image. Note: We set $α = 0.8$, $β1$, $β2$ and $β3 ∼ U(−1, 1)$.

### A. Implementation –

TensorFlow is used to deploy our GAN methodology. [12] A discriminator is suggested to check whether the result and associated cartoon pictures have comparable surfaces and to control the generator to learn the information encoded in the extracted surface representation. To separate the regions, we use the Felzenszwalb method. We employ a pretrained VGGNetwork [13] to enforce a spatial limitation on content between outputs and given matching cartoons.

### B. Dataset –

Data on human faces and landscapes are collected for standardization across different situations. The data contains both real-world and cartoon pictures, but the test data only comprises real-world photographs. All of the photos have been scaled and cropped to $256*256$ pixels. Photos are obtained from the Internet and used for testing.

### C. Learning Rate –

When fine-tuning our model's hyperparameters, we initially used a grid search to obtain an ideal learning rate of *0.001*. Our mini-batch size was restricted to two because we were testing locally due to GCloud resource limitations.

### D. Experimental Result –

As we can see in the cartoonized pictures below, they are very similar when it comes to the sharpness of the object and the presence of different colours in the pictures. Additionally, elements like reflection, shadows are depicted with precision.

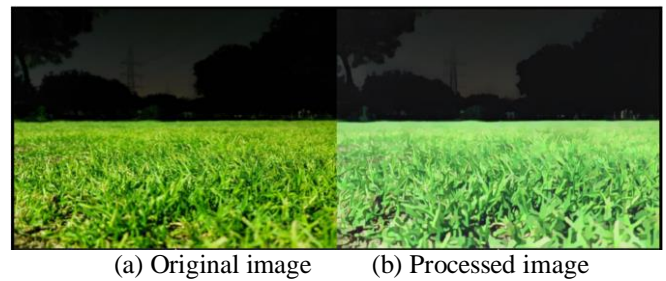

(a) Original image      (b) Processed image

Fig. (6)

### E. Qualitative Evaluation –

We offer findings from qualitative tests with information from four distinct methodologies and original photos, as well as qualitative analysis. We use Frechet Inception Distance (FID) [14] to evaluate performance in quantitative studies by measuring the distance between the source and target image distributions. Candidates in the user study are asked to rate the cartoon quality and overall quality of various techniques on a scale of 1 to 5. Higher scores indicate higher quality.

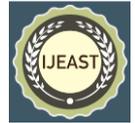

**F. Performance Analysis –**

     Our model is the quickest of the four techniques on all devices and resolutions, and it has the shortest model size. Our model, in particular, can process a *720*1280* picture on GPU in only *17.23ms*, allowing it to do real-time High-Resolution video processing workloads. Generality to a wide range of use cases: we test our model on a variety of real-world situations, such as natural landscapes, city vistas, people, and animals.

## V. CONCLUSION

     In this paper, we proposed a deployed white-box controllable image cartoonization framework based on GAN, which can generate high-quality cartoonized images from real-world photos. Images are decomposed into three cartoon representations: the surface representation, the structure representation, and the texture representation. Corresponding image processing modules are used to extract three representations for network training. Extensive quantitative and qualitative experiments have been conducted to validate the performance of our method.

## VI. FUTURE WORK

     Meanwhile, existing systems cannot produce satisfactory results in terms of cartoonization. But further research in this field can lead to various applications in other domains. The model could help generate quick prototypes or sprites for anime, cartoons and games. Games can import shortcut scenes very easily without using motion capture. Also, since it subdues facial features and information in general, it can be used to generate minimal art. It can be modelled as an assistant to graphic designers or animators so the artists can use it to design and produce unique artworks.